\documentclass[preprint,12pt,authoryear]{elsarticle}

\usepackage{amssymb}
\usepackage{amsmath}

\usepackage{hyperref}
\usepackage{float}
\usepackage{verbatim} 
\usepackage{apalike}
\usepackage{xcolor}
\usepackage{graphicx}
\usepackage{subcaption}
\usepackage{array}
\usepackage{amssymb}
\usepackage{amsmath}

\journal{Nuclear Physics B}

\begin{document}

\begin{frontmatter}



\title{DeepHistoViT: An Interpretable Vision Transformer Framework for Histopathological Cancer Classification} 



\author[label1]{Ravi Mosalpuri}

\author[label1]{Mohammed Abdelsamea}

\author[label1,label2]{Ahmed Karam Eldaly\corref{cor1}}
\ead{a.karam-eldaly@exeter.ac.uk}

\cortext[cor1]{Corresponding author}

\affiliation[label1]{%
    organization={Department of Computer Science, University of Exeter},
    addressline={Innovation Centre, Rennes Drive},
    city={Exeter},
    postcode={EX4 4QF},
    state={Devon},
    country={United Kingdom}
}

\affiliation[label2]{%
    organization={UCL Hawkes Institute, University College London},
    addressline={Gower Street},
    city={London},
    postcode={WC1E 6AE},
    country={United Kingdom}
}

\begin{abstract}
Histopathology remains the gold standard for definitive cancer diagnosis due to its ability to provide detailed cellular-level assessment of tissue morphology. However, manual histopathological examination is time-consuming, labour-intensive, and subject to inter-observer variability, creating a growing demand for reliable computer-assisted diagnostic tools. Recent advances in deep learning, particularly transformer-based architectures, have demonstrated strong potential for modelling complex spatial dependencies in medical images. In this work, we propose DeepHistoViT, a transformer-based deep learning framework for automated classification of histopathological images. The proposed model employs a customized Vision Transformer (ViT) architecture with an integrated attention mechanism designed to capture fine-grained cellular structures while enhancing interpretability through attention-based localization of diagnostically relevant regions. The framework is evaluated on three publicly available and diverse histopathology datasets covering lung cancer, colon cancer, and acute lymphoblastic leukaemia. Experimental results demonstrate that DeepHistoViT achieves state-of-the-art performance across all datasets, with classification accuracy, precision, recall, F1 score, and ROC–AUC reaching 100\% on the lung and colon cancer datasets, and 99.85\%, 99.84\%, 99.86\%, 99.85\%, and 99.99\%, respectively, on the acute lymphoblastic leukaemia dataset. All performance metrics are reported with 95\% confidence intervals to ensure statistical robustness. These findings demonstrate the effectiveness of transformer-based architectures for histopathological image analysis and highlight the potential of DeepHistoViT as a reliable and interpretable computer-assisted diagnostic tool to support pathologists in clinical decision-making. The proposed framework may contribute to improved diagnostic efficiency, consistency, and early cancer detection.

\end{abstract}

\begin{graphicalabstract}
\centering
\includegraphics[width=\textwidth]{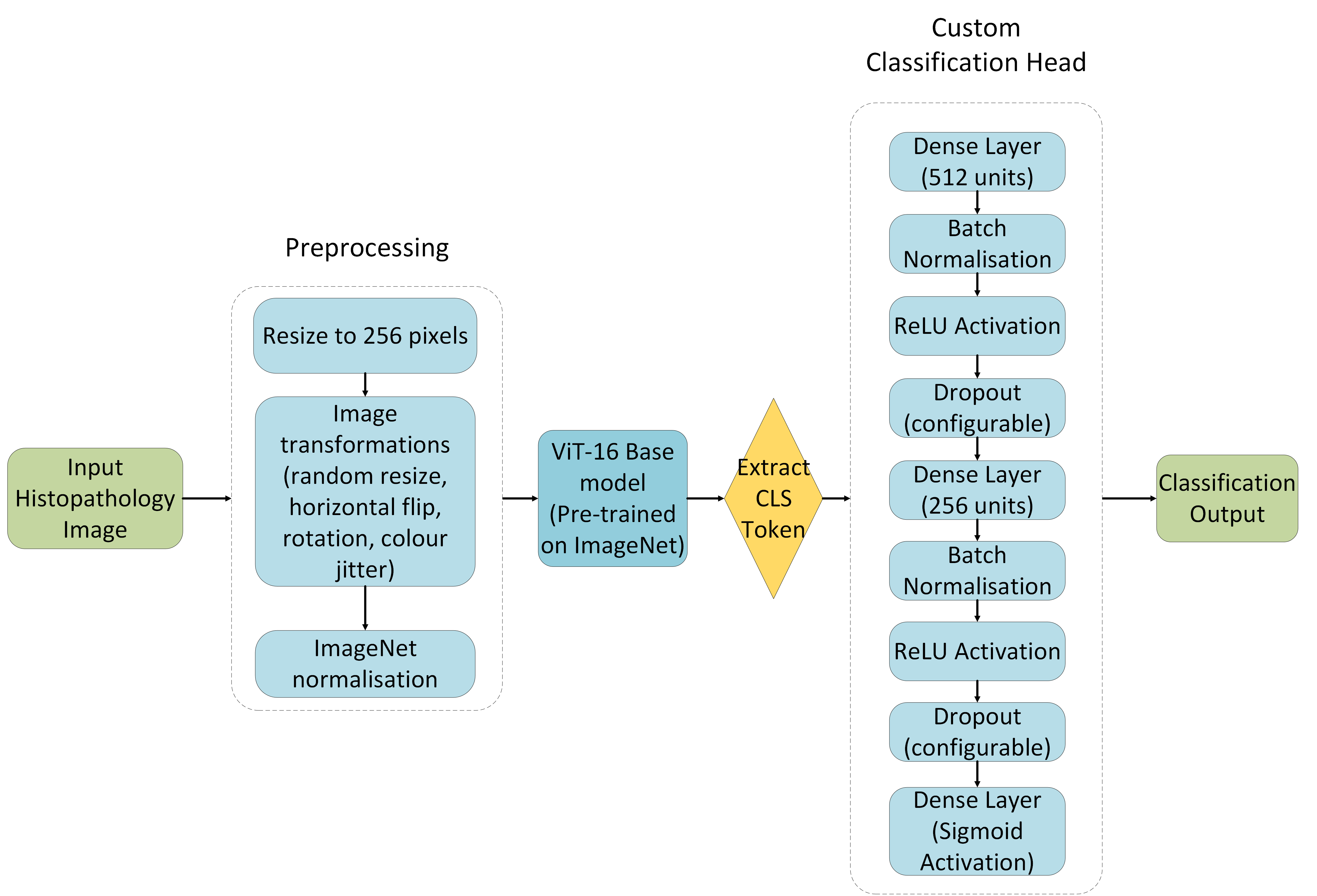}\\
\end{graphicalabstract}

\begin{highlights}

\item DeepHistoViT: Vision Transformer for multi-cancer histopathology classification

\item Achieves 100\% accuracy on LC25000 and 99.85\% on ALL leukemia dataset

\item Custom classification head improves feature adaptation and generalisation

\item Attention maps provide interpretable localisation of diagnostic regions

\item Robust performance without stain normalisation or domain harmonisation

\end{highlights}

\begin{keyword}


Histopathology \sep Cancer \sep Deep learning \sep Image Classification \sep Vision Transformers
\end{keyword}

\end{frontmatter}

\section{Introduction}
\label{introduction}

Cancer is identified by the world health organisation (WHO) as a leading cause of death worldwide, responsible for approximately 10 million deaths in 2020 \cite{WHO2025CancerFactSheet}. The WHO reported that in 2021, non-communicable diseases constituted nearly 75\% of all non-pandemic-related deaths globally, with cancer being one of the primary causes \cite{WHO2025NCDFactSheet}. Cancer is characterised by the uncontrolled growth of abnormal cells within the body, which can metastasise to other organs, making this the primary factor in cancer-related deaths. Several factors contribute to the formation of cancerous cells, including genetic mutations, lifestyle choices (such as smoking, unhealthy diets, and a lack of physical activity), and environmental factors (such as exposure to chemicals or radiation). Accurate detection of cancer is crucial in determining appropriate treatment strategies, as each type has unique biological characteristics and treatment requirements.

The GLOBOCAN 2020 report \cite{GLOBOCAN2020} highlights the significant impact of several major cancer types worldwide. The breast, lung, prostate, and colorectal cancers are among the most prevalent, collectively accounting for a large proportion of new cancer cases and cancer-related deaths each year. In 2020, breast cancer led with over 2.3 million new cases globally, followed by lung, colorectal, and prostate cancers, each contributing substantially to the global cancer burden. Survival rates and symptoms vary extensively between cancer types. The American Cancer Society’s 2025 report \cite{ACS2025CancerFactsFigures} indicates that the overall 5-year relative survival rate is 27\% for lung cancer, 65\% for colorectal cancer, and can exceed 90\% for localised breast cancer cases when detected early. Common warning signs can range from persistent cough and chest discomfort in lung cancer, to abnormal bleeding or lumps in the breast and colorectal cancers. However, these symptoms are typically only apparent when the cancer has reached an advanced stage \cite{ACS2025CancerFactsFigures}. Therefore, early and accurate detection of cancer through histopathology images can drastically improve survival chances by prompt treatment.

Histopathology remains the gold standard for definitive cancer diagnosis, as it enables detailed microscopic examination of tissue morphology and cellular architecture by trained pathologists. Manual assessment of histopathological slides plays a central role in clinical decision-making; however, this process is labour-intensive, time-consuming, and subject to inter- and intra-observer variability, particularly when large volumes of whole-slide images must be analysed \cite{Campanella2019NatureMed, litjens2016deep}. Traditional computer-aided diagnosis approaches based on handcrafted features and classical image processing techniques have been explored to support pathologists, but these methods often require extensive manual feature engineering and lack robustness when applied to large-scale, heterogeneous datasets \cite{Gurcan2009Review, Irshad2014Review, Litjens2017MedImageAnalysis, Komura2018Review, eldaly2015bag, eldaly2019patch, eldaly2019bayesian}. In contrast, deep learning has transformed digital pathology by enabling automated feature extraction directly from raw image data, eliminating the need for manual feature design and improving scalability and diagnostic performance \cite{Litjens2017MedImageAnalysis, Campanella2019NatureMed, Echle2021LancetOncology, Lu2021NatMed, Steiner2020NatMed, tizhoosh2018artificial}. Convolutional neural networks (CNNs), in particular, have demonstrated strong capability in learning hierarchical spatial features and have achieved expert-level performance in various histopathological image classification tasks, including cancer detection and subtype classification \cite{Esteva2017Nature, Coudray2018NatMed, Campanella2019NatureMed, Steiner2020NatMed, Echle2021LancetOncology, Wang2016TMI}. They automatically extract features from raw images through a series of convolutions, pooling layers and non-linear activation functions. CNNs are capable of identifying more intricate patterns and structures in images that are appropriate for medical imaging applications. Additionally, DL algorithms can process large volumes of digital pathology images with speed and reproducibility, thereby reducing the time required for manual examination.

Although CNNs excel at feature extraction, they struggle in capturing the spatial relationships among different features. It is essential for algorithms to understand global contextual information, as the spatial relationships among tissue structures can greatly impact classification results. Considering the convolution architecture struggles to comprehend global information, researchers have proposed various architectural modifications. With the increasing popularity of transformers in natural language processing (NLP), a variant of the transformer architecture was introduced, namely the vision transformer (ViT). The transformer architecture can handle a series of fixed-size image patches as input, which helps in extracting complex features from images. It uses a global attention mechanism to tackle long-range dependency issues often linked with CNNs. ViTs have led to significant progress in medical computer vision across different imaging methods \cite{Hayat2025, YAN2025}. In particular, within the field of histopathological image classification, ViTs have demonstrated exceptional accuracy in diagnosing cancers, including breast, lung, and colon cancer \cite{Hayat2025, YAN2025, Shahadat2025}.

Despite the recent advancements in DL for the classification of cancer, significant research gaps remain that hinder the effective implementation of these models in clinical environments. A primary concern is the generalisability of these models, for instance, models that are trained only on augmented datasets like LC25000 \cite{LC25000} usually perform well with similar datasets only and tend to face challenges when dealing with real-world clinical images. On the other hand,iInterpretability is another key concern, as most DL models function as opaque ‘black-boxes’. Therefore, establishing clinician trust is essential for widespread adoption. There is still limited adoption of Explainable Artificial Intelligence (XAI) algorithms, popularly known as XAI techniques.

This work introduces DeepHistoVit, a customised ViT-16–based deep learning framework specifically fine-tuned for the classification of histopathology images across multiple cancer types. The main contributions of this work are as follows.

\begin{enumerate}
    \item We introduce \textit{DeepHistoVit}, a customised Vision Transformer (ViT-16) model, specifically designed for the classification of histopathology images across three diverse cancer datasets, including lung, acute lymphoblastic leukaemia, and colon. By incorporating a tailored multi-layer classification head with batch normalisation, configurable dropout, and progressive dimensionality reduction, the model effectively captures complex tissue features while remaining broadly adaptable across heterogeneous datasets. Notably, the framework achieves this without requiring stain normalisation, demonstrating practical applicability to real-world histopathology data.

    \item DeepHistoVit integrates an attention-based visualisation mechanism that highlights the most informative regions of each histopathology image. This feature enhances interpretability and allows pathologists to inspect the model's reasoning at the patch level, providing a bridge between computational predictions and clinical validation, and fostering trust in AI-assisted diagnostic tools.

    \item We rigorously assess the performance of DeepHistoVit across three independent datasets, reporting metrics including accuracy, precision, recall, F1 score, and ROC–AUC along with 95\% confidence intervals. The results demonstrate consistent, state-of-the-art performance and confirm the model's ability to generalise across distinct tissue types, highlighting its potential as a reliable and versatile tool for digital pathology.
\end{enumerate}

The remaining sections of the paper are organised as follows. Section \ref{sec:Literature Review} reviews the literature for histopathology image classification using deep learning. The design and implementation details of the proposed approach are presented in Section \ref{sec:Methodology}. Experimental results are then presented in Section \ref{sec:Results}. Discussions and general insights into the proposed approach, including main limitations and plans for future work are discussed in Section \ref{sec:Discussion}. Finally, conclusions are outlined in Section \ref{sec:Conclusions}.

\section{Related Work}
\label{sec:Literature Review}
Recent advances in deep learning have significantly improved the automated classification of histopathological images, enabling more accurate and efficient cancer diagnosis. Early approaches primarily relied on convolutional neural networks (CNNs), which demonstrated strong capability in learning hierarchical spatial representations from microscopic tissue images. For example, Shahadat et al.~\cite{Shahadat2025} proposed a lightweight CNN incorporating squeeze-and-excitation modules for lung and colon cancer classification using the LC25000 dataset~\cite{LC25000}. Their architecture achieved 100\% accuracy while maintaining a low computational footprint, highlighting the feasibility of efficient CNN-based models for clinical deployment. Similarly, Bala et al.~\cite{Bala2024} introduced MRANet, a residual attention-based CNN that integrates attention modules to enhance feature representation, achieving accuracies up to 99.30\% in lung cancer classification tasks. These studies demonstrate the effectiveness of attention-enhanced CNN architectures in capturing discriminative histopathological features.

To further improve classification performance and robustness, several works have explored multi-model fusion and feature selection strategies. Attallah et al.~\cite{Attallah2025} developed a computer-aided diagnostic framework combining features extracted from MobileNet, EfficientNetB0, and ResNet-18 architectures. By applying canonical correlation analysis and statistical feature selection methods, their approach achieved 99.78\% classification accuracy on the LC25000 dataset. Ensemble learning approaches have also demonstrated strong performance. Atwan et al.~\cite{Atwan2024} proposed an ensemble framework combining ResNet50 and MobileNet feature extractors with classical classifiers, achieving 99.2\% and 97.72\% accuracy for binary and multiclass breast cancer classification, respectively. These hybrid and ensemble approaches improve robustness by leveraging complementary feature representations from multiple architectures, although they often increase architectural complexity.

Transfer learning has also emerged as an effective strategy for histopathological image classification by leveraging pretrained models to improve generalisation. Ochoa-Ornelas et al.~\cite{Ochoa-Ornelas2025} fine-tuned EfficientNetB3 using both LC25000 and clinically sourced genomic data commons images, achieving 99.39\% accuracy while improving clinical realism and generalisability. Similarly, El-Aziz et al.~\cite{ElAziz2025} introduced a learning rate tuning strategy for EfficientNet-B3, enabling dynamic adaptation during training and achieving 99.84\% accuracy for oral cancer classification. These approaches demonstrate the effectiveness of transfer learning and training optimisation techniques in improving model performance and convergence.

Despite the success of CNN-based methods, their inherently local receptive fields limit their ability to capture long-range dependencies and global contextual relationships within histopathological images. To address this limitation, transformer-based architectures have recently gained increasing attention due to their ability to model global feature interactions through self-attention mechanisms. Hayat et al.~\cite{Hayat2025} proposed a hybrid EfficientNetV2 and Vision Transformer framework, achieving 99.83\% accuracy on binary breast cancer classification tasks. Similarly, Yan et al.~\cite{YAN2025} introduced DWNAT-Net, a transformer-based architecture incorporating discrete wavelet transforms to jointly capture spatial and frequency-domain features, achieving 99.66\% accuracy on the BreakHis dataset and demonstrating strong capability in modelling complex histopathological structures.

In addition to architectural advancements, human-in-the-loop learning frameworks have been explored to enhance model reliability and performance. Han et al.~\cite{han2025towards} proposed an iterative training framework incorporating expert feedback to refine model predictions, demonstrating improved accuracy and robustness in invasive ductal carcinoma detection. While such approaches improve reliability, their reliance on continuous expert intervention limits scalability in routine clinical practice.

Although existing methods have achieved promising results, several challenges remain. CNN-based models are limited in capturing global contextual dependencies, while hybrid and ensemble approaches often introduce increased computational complexity and architectural redundancy. Furthermore, interpretability remains a critical challenge, as many existing models provide limited transparency regarding the regions influencing model predictions. Transformer-based architectures offer a promising solution due to their inherent ability to capture long-range dependencies while providing attention-based interpretability.

To address these limitations, this work proposes DeepHistoViT, a customised Vision Transformer framework specifically designed for histopathological image classification. The proposed model leverages self-attention mechanisms to capture both fine-grained cellular structures and global contextual relationships, while providing interpretable attention maps highlighting diagnostically relevant regions. Unlike hybrid or ensemble approaches, DeepHistoViT provides a unified transformer-based framework capable of achieving state-of-the-art performance across multiple cancer types while maintaining architectural simplicity, scalability, and interpretability.


\section{Methodology}
\label{sec:Methodology}

\subsection{Datasets}

The performance of the proposed DeepHistoViT framework is evaluated using three publicly available histopathology image datasets representing different cancer types. These datasets provide diverse morphological characteristics and imaging conditions, enabling comprehensive evaluation of model robustness and generalisation.

\begin{enumerate}
\item {Colon Cancer Dataset (LC25000)}  
Colon cancer images are also obtained from the LC25000 dataset~\cite{LC25000}. Two classes are considered: colon adenocarcinoma and benign colon tissue. Although derived from the same dataset as lung cancer, colon classification is treated as a separate task due to distinct morphological characteristics and classification objectives.

\item {Lung Cancer Dataset (LC25000)}  
The lung cancer images are obtained from the LC25000 dataset~\cite{LC25000}, which contains 25,000 histopathological images of lung and colon tissues. For lung cancer classification, three classes are used: lung adenocarcinoma, lung squamous cell carcinoma, and benign lung tissue. Each class contains 5,000 RGB images with a resolution of $768 \times 768$ pixels. All images are anonymised and compliant with the Health Insurance Portability and Accountability Act (HIPAA), ensuring patient privacy and ethical use. This dataset provides high-quality cellular structures suitable for automated cancer classification.

\item {Acute Lymphoblastic Leukaemia Dataset (ALL)}  
The Acute Lymphoblastic Leukaemia dataset~\cite{ALLDataset} contains 3,562 peripheral blood smear images collected from 89 patients. The dataset includes benign hematogone cells and malignant lymphoblast cells, which comprise Early Pre-B, Pre-B, and Pro-B subtypes. These images exhibit significant morphological variability, providing a challenging benchmark for classification models.
\end{enumerate}

\begin{figure}[t]
\centering
\includegraphics[width=0.8\linewidth]{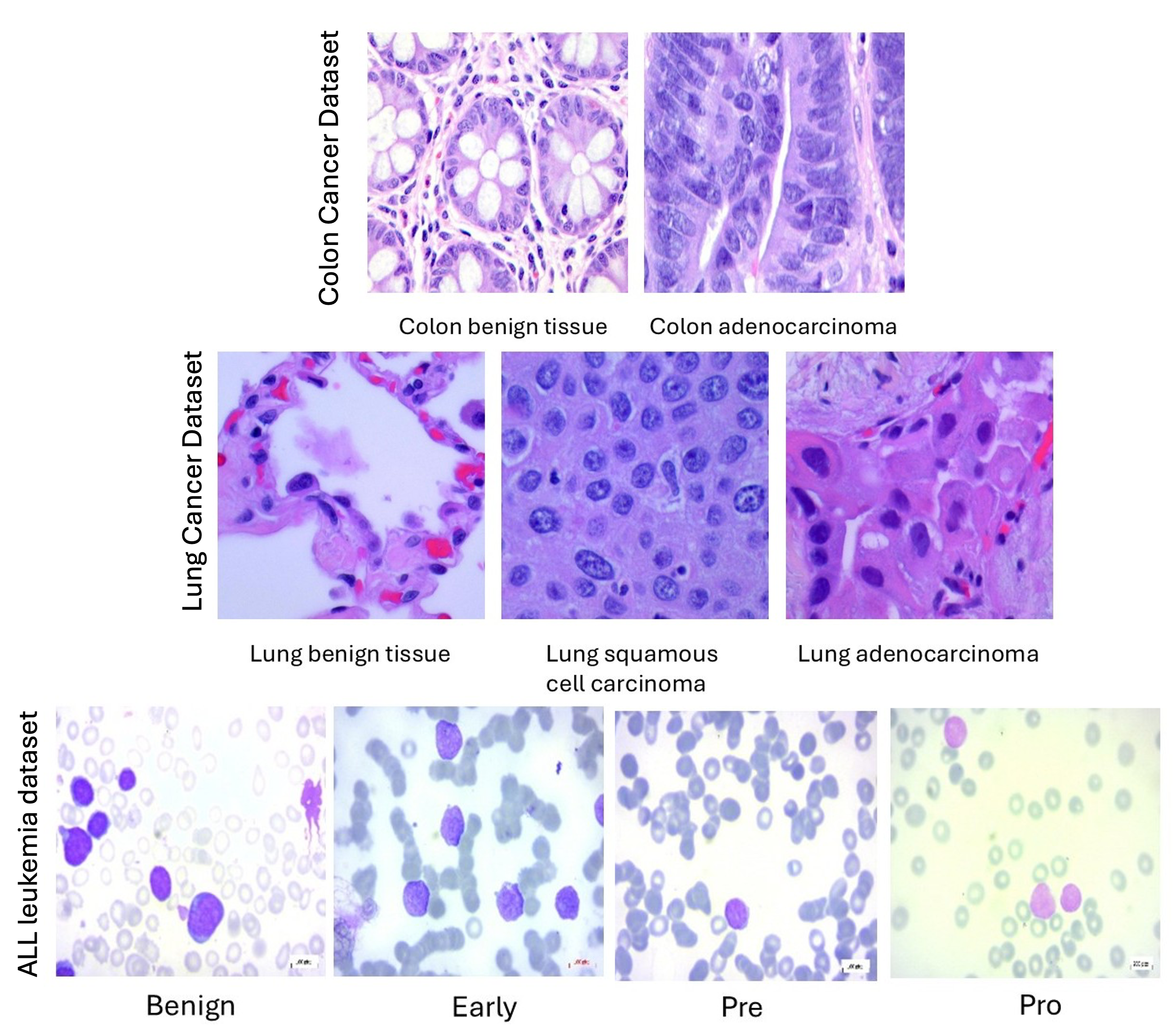}
\caption{Example histopathological images from the datasets used in this study. 
{Top row:} Colon tissue samples from the LC25000 dataset showing (h) benign colon tissue and (i) colon adenocarcinoma. {Middle row:} Lung tissue samples showing (a) benign lung tissue, (b) squamous cell carcinoma, and (c) adenocarcinoma. {Bottom row:} Peripheral blood smear images from the acute lymphoblastic leukaemia (ALL) dataset showing (d) benign cells, (e) early Pre-B lymphoblasts, (f) Pre-B lymphoblasts, and (g) Pro-B lymphoblasts. All histopathological images are stained using hematoxylin and eosin (H\&E).}
\label{fig:pic1}
\end{figure}

Representative examples from each dataset are shown in Figures~\ref{fig:pic1}.

\subsection{Data Preparation and Preprocessing}
\label{subsec:preprocessing}
All histopathology images are subjected to a standardised preprocessing pipeline to ensure compatibility with the Vision Transformer architecture while preserving diagnostically relevant morphological information. Specifically, each image is first resized to $256 \times 256$ pixels and subsequently centre-cropped to $224 \times 224$ pixels. This spatial resolution aligns with the input dimensional requirements of the pretrained ViT-base-patch16-224 model~\cite{dosovitskiy2021ViT}, and ensures consistent patch extraction. This step defines the effective image domain from which fixed-size patches are generated and embedded into the transformer input sequence.

\begin{figure}
    \centering
    \includegraphics[width=\linewidth]{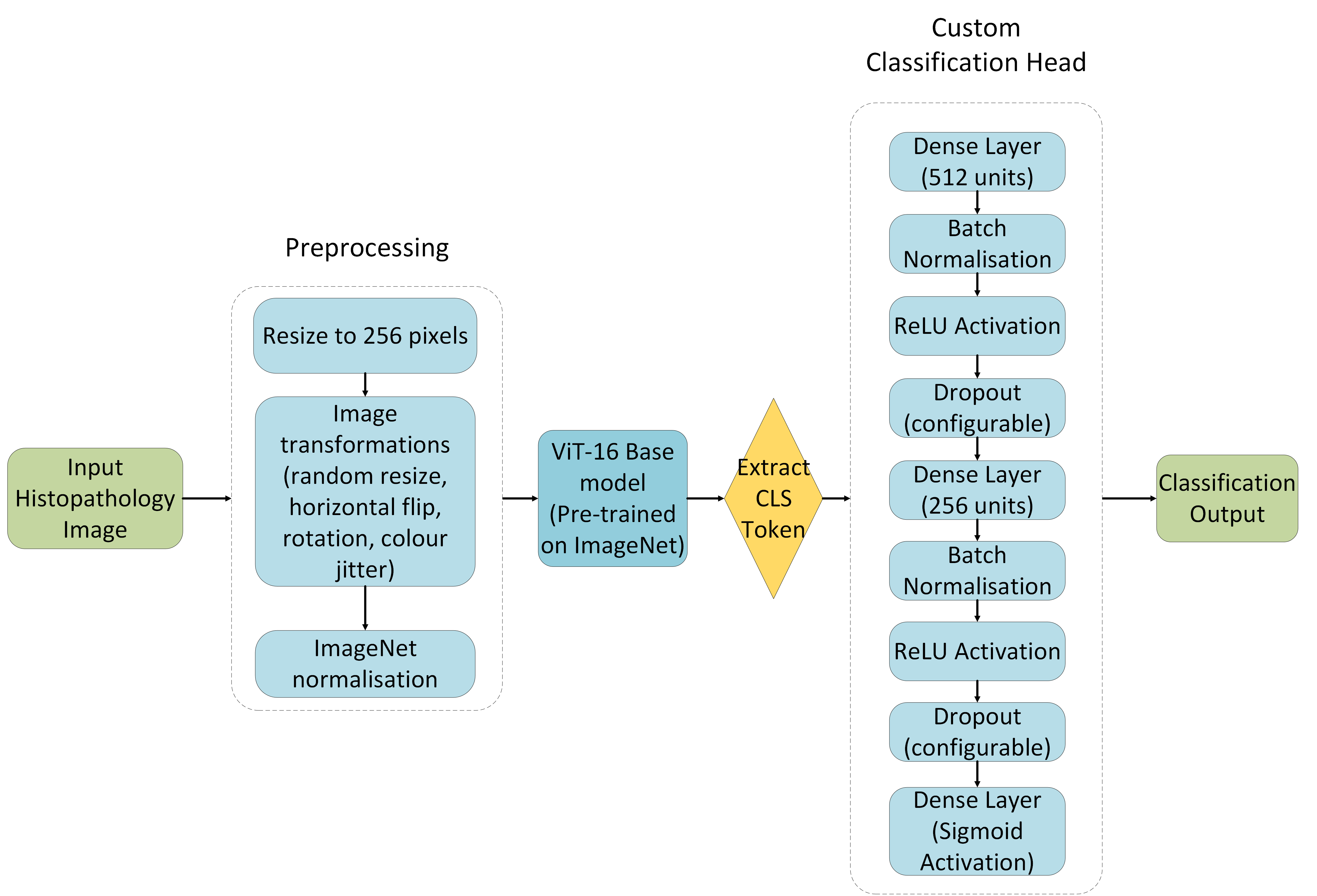}
    \caption{Overview of the proposed Vision Transformer (ViT‑16) based pipeline for histopathology image classification. The framework consists of three stages: (i) Preprocessing, where raw histopathology images are resized to 256 pixels, augmented through random resizing, horizontal flipping, rotation, and colour jitter, followed by ImageNet normalisation; (ii) Feature Extraction, leveraging a ViT‑16 base model pre‑trained on ImageNet to obtain the CLS token representation; and (iii) Custom Classification Head, comprising sequential dense layers (512 and 256 units) with batch normalization, ReLU activation, and configurable dropout, culminating in a sigmoid‑activated output layer. This architecture integrates transformer‑based global feature learning with a tailored classification head to enhance performance on medical image analysis tasks.}
    \label{fig:proposed model}
\end{figure}

No stain normalisation or domain harmonisation techniques are applied. This decision preserves the intrinsic variability present in histopathology images, including differences in staining intensity, scanner characteristics, and tissue preparation protocols. By retaining natural variability, the proposed framework is evaluated under conditions that more closely reflect real-world clinical deployment. To improve model generalisation and mitigate overfitting, data augmentation is applied during training. The augmentation pipeline includes random resized cropping, horizontal flipping, rotations up to $20^\circ$, and colour jittering. These transformations simulate realistic acquisition variability and encourage the model to learn invariant morphological representations. Validation and test sets undergo only deterministic preprocessing consisting of resizing and centre cropping, ensuring unbiased evaluation. Following preprocessing, all images are normalised using the channel-wise mean and standard deviation of the ImageNet dataset~\cite{ImageNet}, with mean values of $[0.485, 0.456, 0.406]$ and standard deviation values of $[0.229, 0.224, 0.225]$. This normalisation aligns the input distribution with that of the pretrained backbone and facilitates stable optimisation during transfer learning. The complete preprocessing and feature extraction pipeline is illustrated in Fig.~\ref{fig:proposed model}.

\subsection{Model Architecture}
\label{subsec:architecture}
The proposed DeepHistoViT framework builds upon the Vision Transformer (ViT) architecture originally introduced by Dosovitskiy et al.~\cite{dosovitskiy2021ViT}. Unlike convolutional neural networks, which rely on local receptive fields, the Vision Transformer models an image as a sequence of embedded patches and processes this sequence using stacked self-attention layers, enabling direct modelling of global contextual dependencies.

Given a preprocessed input image $x \in \mathbb{R}^{H \times W \times C}$, the image is divided into $N$ non-overlapping patches of size $P \times P$. Each patch is flattened and projected into a $D$-dimensional embedding space using a learnable linear projection. Positional embeddings are added to retain spatial information that would otherwise be lost during patch flattening. A learnable classification token is prepended to the sequence to aggregate global information across all patches. The resulting sequence is processed through multiple transformer encoder layers. Each encoder layer consists of a multi-head self-attention module followed by a feed-forward network. The attention mechanism computes Query ($Q$), Key ($K$), and Value ($V$) matrices, and the attention output is given by

\begin{equation}
\text{Attention}(Q,K,V) =
\text{Softmax}\left(\frac{QK^T}{\sqrt{d_k}}\right)V,
\label{eq:attention}
\end{equation}
where $d_k$ denotes the dimensionality of the key vectors. This mechanism enables the model to learn global relationships between spatially distant regions, which is particularly important for histopathological image analysis, where diagnostic features may be distributed across the tissue.

The backbone network is initialised using pretrained weights from the ViT-base-patch16-224 model trained on the ImageNet dataset~\cite{ImageNet, dosovitskiy2021ViT}. Transfer learning allows the model to leverage general visual representations learned from large-scale natural image datasets and adapt them to histopathological image classification. To improve adaptation to the target domain, a selective fine-tuning strategy is employed, where only the final two transformer encoder blocks are unfrozen during training. This enables domain-specific feature refinement while preserving general low-level representations. Furthermore, a custom multi-layer classification head is introduced, replacing the default linear classifier. As shown in Fig.~\ref{fig:proposed model}, the classification token embedding is passed through two fully connected layers with progressive dimensionality reduction (768$\rightarrow$512$\rightarrow$256), each followed by batch normalisation, ReLU activation, and dropout. This structure enables progressive feature refinement and improves discrimination between histopathological tissue classes. The final classification layer maps the refined feature representation to the output space, producing class probabilities via the softmax function.


\subsection{Model Training}
\label{subsec:model_training}
The proposed DeepHistoViT framework is implemented using the PyTorch deep learning library and trained using GPU acceleration. Training is conducted on high-performance computing infrastructure, including NVIDIA Tesla T4 and P100 GPUs, to ensure efficient optimisation of the transformer-based architecture. To ensure reproducibility, all experiments are conducted with fixed random seeds across Python, NumPy, and PyTorch, including CUDA operations. This ensures consistent weight initialisation, data splitting, and training behaviour across repeated runs.

Each dataset is partitioned into training, validation, and test sets using stratified sampling to preserve class distributions. Specifically, 20\% of the data is reserved as an independent test set, which remains completely unseen during training and hyperparameter optimisation. The remaining 80\% is further divided into training and validation sets using an 80:20 ratio. This results in an effective split of 64\%, 16\%, and 20\% for training, validation, and testing, respectively. Model training is performed by minimising the categorical cross-entropy loss function.

Model optimisation is performed using the Adaptive Moment Estimation (Adam) optimiser~\cite{AdamOptim2014}, which computes adaptive learning rates for each parameter based on first and second moments of the gradients. Hyperparameter optimisation is conducted using grid search over key parameters, including learning rate, weight decay, dropout rate, and batch size. The search space consists of learning rates $\{10^{-4}, 10^{-3}, 10^{-2}\}$, weight decay values $\{10^{-5}, 10^{-4}, 10^{-3}\}$, dropout rates $\{0.2, 0.3, 0.4\}$, and batch sizes $\{16, 32, 64\}$. The optimal hyperparameter configuration is selected based on validation accuracy. To improve convergence stability and prevent overfitting, a ReduceLROnPlateau learning rate scheduler is employed, which reduces the learning rate when validation performance plateaus. In addition, early stopping is used with a patience of eight epochs, terminating training when validation performance no longer improves. The model parameters corresponding to the best validation performance are retained for final evaluation.

\section{Experimental Results}
\label{sec:Results}

This section presents a comprehensive evaluation of the proposed DeepHistoViT framework across three histopathology datasets. The experimental analysis focuses on hyperparameter optimisation, model generalisation, and classification performance using multiple quantitative metrics. The evaluation protocol is designed to ensure statistical reliability, reproducibility, and unbiased assessment of model performance.

\subsection{Hyperparameter Optimisation}
\label{subsec:hyperparameter}
Hyperparameter optimisation is performed using a systematic grid search over key training parameters, including learning rate, weight decay, dropout rate, and batch size, as described in Section~\ref{sec:Methodology}. The objective is to identify parameter configurations that maximise validation performance while maintaining stable convergence and minimising overfitting. The optimal hyperparameter configuration across datasets consists of a weight decay of $1 \times 10^{-5}$, dropout rate of 0.2, and batch size of 16 for the lung and colon cancer datasets, and 32 for the leukaemia dataset. These values provide an effective balance between regularisation and model capacity. The selected weight decay introduces sufficient regularisation to prevent overfitting while preserving discriminative feature learning. Similarly, the dropout rate of 0.2 improves generalisation by reducing co-adaptation of neurons within the classification head. The batch size selection reflects dataset-specific optimisation, with the larger batch size for the leukaemia dataset facilitating more stable gradient estimation due to its smaller dataset size. The learning rate is selected based on validation performance to ensure stable optimisation of the fine-tuned transformer layers while preserving pretrained feature representations.

\subsection{Evaluation Protocol}
\label{subsec:evaluation_protocol}
To ensure robust and unbiased performance estimation, a stratified five-fold cross-validation strategy is employed on the training data. This approach preserves class distribution across folds and enables evaluation of model stability under different data partitions. Cross-validation performance is used for model selection and hyperparameter optimisation. In addition to cross-validation, a strictly independent held-out test set, comprising 20\% of the dataset as defined in Section~\ref{sec:Methodology}, is used exclusively for final performance evaluation. This separation ensures that test data are not used during model training or hyperparameter selection, thereby preventing data leakage and providing an unbiased estimate of generalisation performance. To further improve prediction robustness, test-time augmentation (TTA) is applied during inference~\cite{TestTimeAug}. Each test image is evaluated under multiple geometric transformations, including horizontal and vertical flips and rotations of $\pm 10^\circ$. The final prediction is obtained by averaging the predicted class probabilities across all augmented versions of the image. This approach reduces sensitivity to image orientation and improves prediction stability.

\subsection{Evaluation Metrics}
\label{subsec:metrics}
Model performance is evaluated using multiple quantitative metrics to provide a comprehensive assessment of classification accuracy and reliability. These metrics include accuracy, precision, recall, F1-score, and area under the receiver operating characteristic curve (ROC-AUC). Accuracy measures the overall proportion of correctly classified samples. Precision quantifies the proportion of correctly predicted positive samples among all predicted positives, while recall measures the proportion of correctly identified positive samples among all true positives. The F1-score provides a balanced measure by combining precision and recall. The ROC-AUC metric evaluates the model’s ability to discriminate between classes across different classification thresholds and is particularly informative for assessing classification robustness. In addition to scalar metrics, confusion matrices are analysed to examine class-specific performance and identify potential misclassification patterns. To ensure statistical reliability, performance metrics are reported on the independent test set and, where applicable, averaged across cross-validation folds.

\subsection{Quantitative Results}
\label{subsec:quantitative_results}

The quantitative performance of the proposed DeepHistoViT framework is evaluated on three independent histopathology datasets, including lung and colon cancer images from the LC25000 dataset~\cite{LC25000}, and acute lymphoblastic leukaemia images from the ALL dataset~\cite{ALLDataset}. Model performance is assessed using accuracy, precision, recall, F1-score, and area under the receiver operating characteristic curve (ROC-AUC), as defined in Section~\ref{subsec:metrics}.

On the lung cancer dataset, the proposed model achieves an overall classification accuracy of 100\% on the held-out test set. As shown in Table~\ref{tab:classification_report_lung}, precision, recall, and F1-score values of 1.0000 are obtained for all classes, including benign lung tissue, adenocarcinoma, and squamous cell carcinoma. These results indicate that the model successfully separates the three lung tissue classes without misclassification under the defined evaluation protocol.

\begin{table}[h!]
\centering
\caption{Classification performance of the proposed DeepHistoViT model on the LC25000 lung cancer dataset. Precision, recall, and F1-score values are reported for each class, along with overall accuracy.}
\begin{tabular}{lcccc}
\hline
\textbf{Class} & \textbf{Precision} & \textbf{Recall} & \textbf{F1-Score} & \textbf{Support} \\
\hline
Benign & 1.0000 & 1.0000 & 1.0000 & 1000 \\
ACC & 1.0000 & 1.0000 & 1.0000 & 1000 \\
SCC & 1.0000 & 1.0000 & 1.0000 & 1000 \\
\hline
\textbf{Accuracy} & & & 1.0000 & 3000 \\
\textbf{Macro Avg} & 1.0000 & 1.0000 & 1.0000 & 3000 \\
\textbf{Weighted Avg} & 1.0000 & 1.0000 & 1.0000 & 3000 \\
\hline
\end{tabular}
\label{tab:classification_report_lung}
\end{table}

Similarly, on the colon cancer dataset, the proposed model achieves 100\% accuracy on the independent test set. Table~\ref{tab:classification_report_colon} shows that precision, recall, and F1-score values are equal to 1.0000 for both benign and malignant colon tissue classes. These results indicate consistent separation between normal and cancerous colon tissue samples within the dataset.

\begin{table}[h!]
\centering
\caption{Classification performance of the proposed DeepHistoViT model on the LC25000 colon cancer dataset. Precision, recall, and F1-score values are reported for each class, along with overall accuracy.}
\begin{tabular}{lcccc}
\hline
\textbf{Class} & \textbf{Precision} & \textbf{Recall} & \textbf{F1-Score} & \textbf{Support} \\
\hline
Benign & 1.0000 & 1.0000 & 1.0000 & 1000 \\
ACC & 1.0000 & 1.0000 & 1.0000 & 1000 \\
\hline
\textbf{Accuracy} & & & 1.0000 & 2000 \\
\textbf{Macro Avg} & 1.0000 & 1.0000 & 1.0000 & 2000 \\
\textbf{Weighted Avg} & 1.0000 & 1.0000 & 1.0000 & 2000 \\
\hline
\end{tabular}
\label{tab:classification_report_colon}
\end{table}

For the acute lymphoblastic leukaemia dataset, the model achieves an overall accuracy of 99.85\%, with macro-averaged precision, recall, and F1-score values of 99.87\%, 99.75\%, and 99.81\%, respectively. The ROC-AUC reaches 99.99\%, indicating strong class separability. As shown in Table~\ref{tab:classification_report_ALL}, near-perfect performance is observed across all four classes. Minor deviations from perfect classification are observed in the benign class, while the malignant subclasses (Early, Pre, and Pro) are classified with perfect or near-perfect accuracy.

\begin{table}[h!]
\centering
\caption{Classification performance of the proposed DeepHistoViT model on the ALL dataset. Precision, recall, and F1-score values are reported for each class, along with overall accuracy.}
\begin{tabular}{lcccc}
\hline
\textbf{Class} & \textbf{Precision} & \textbf{Recall} & \textbf{F1-Score} & \textbf{Support} \\
\hline
Benign & 1.0000 & 0.9901 & 0.9950 & 101 \\
Early & 0.9949 & 1.0000 & 0.9975 & 197 \\
Pre   & 1.0000 & 1.0000 & 1.0000 & 193 \\
Pro   & 1.0000 & 1.0000 & 1.0000 & 161 \\
\hline
\textbf{Accuracy}     &        &        & 0.9985 & 652 \\
\textbf{Macro Avg}    & 0.9987 & 0.9975 & 0.9981 & 652 \\
\textbf{Weighted Avg} & 0.9985 & 0.9985 & 0.9985 & 652 \\
\hline
\end{tabular}
\label{tab:classification_report_ALL}
\end{table}

\begin{figure}
    \centering
    \begin{subfigure}[b]{0.32\textwidth}
        \centering
        \includegraphics[width=\textwidth]{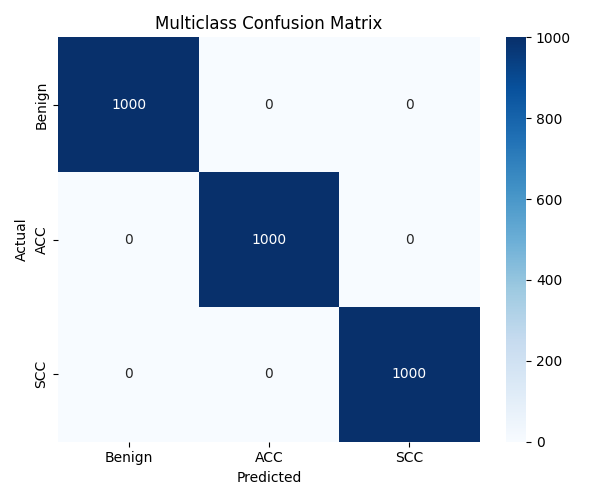}
        \caption{Lung cancer (LC25000)}
        \label{fig:confusion_matrix_lung}
    \end{subfigure}
    \hfill
    \begin{subfigure}[b]{0.32\textwidth}
        \centering
        \includegraphics[width=\textwidth]{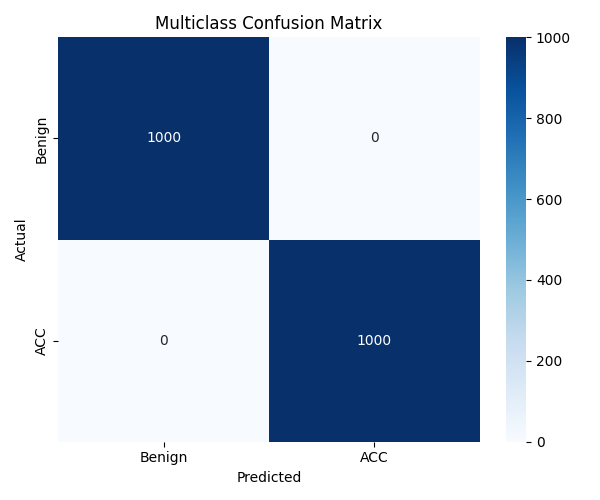}
        \caption{Colon cancer (LC25000)}
        \label{fig:confusion_matrix_colon}
    \end{subfigure}
    \hfill
    \begin{subfigure}[b]{0.32\textwidth}
        \centering
        \includegraphics[width=\textwidth]{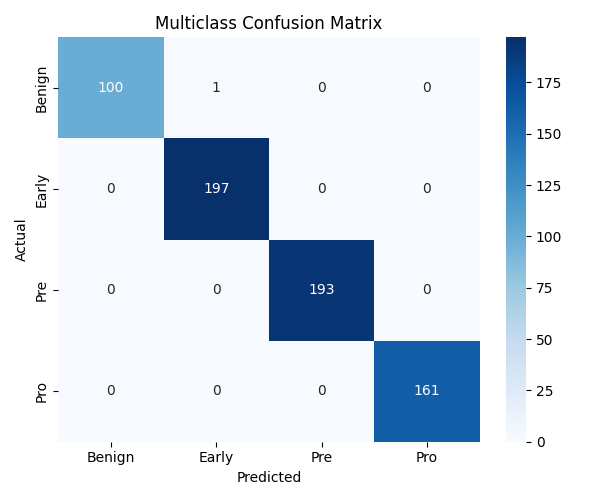}
        \caption{ALL dataset}
        \label{fig:confusion_matrix_all}
    \end{subfigure}
    \caption{Confusion matrices for the three evaluated datasets: lung cancer (LC25000), colon cancer (LC25000), and acute lymphoblastic leukaemia (ALL). The lung and colon datasets show perfect classification performance, while the ALL dataset shows near-perfect classification with minimal misclassification.}
    \label{fig:confusion_matrices_all}
\end{figure}

The confusion matrices in Fig.~\ref{fig:confusion_matrices_all} provide a detailed view of class-wise predictions. For the lung and colon cancer datasets, the confusion matrices show perfect diagonal structure, confirming that all samples are correctly classified. For the ALL dataset, only a small number of misclassifications are observed, primarily involving benign samples, while malignant subclasses are classified correctly with high reliability.

\begin{figure}[h!]
    \centering
    \begin{subfigure}[b]{0.3\textwidth}
        \centering
        \includegraphics[width=\textwidth]{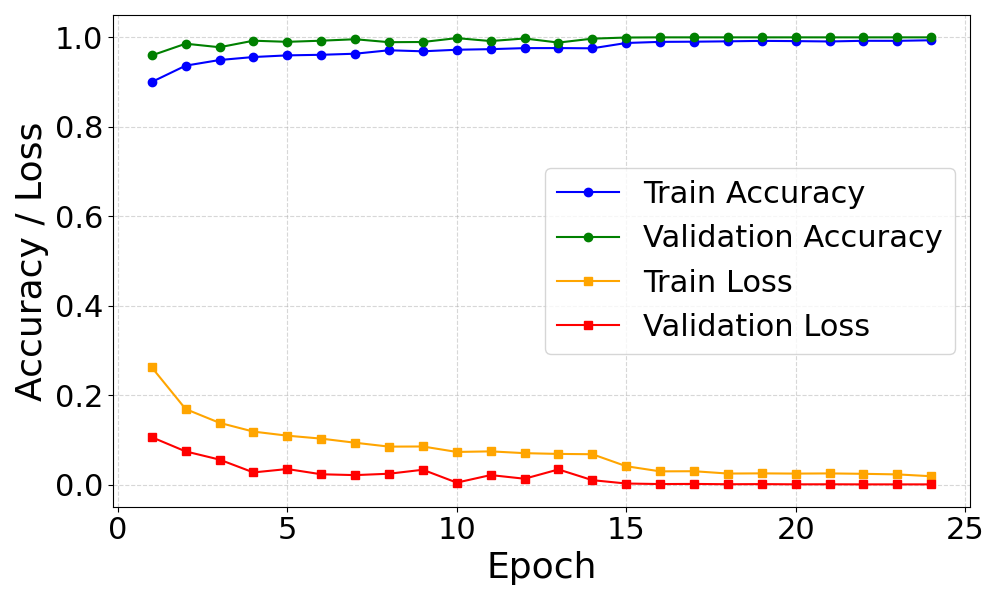}
        \caption{Lung cancer dataset}
        \label{fig:graph_AccVsLoss_lung}
    \end{subfigure}
    \begin{subfigure}[b]{0.3\textwidth}
        \centering
        \includegraphics[width=\textwidth]{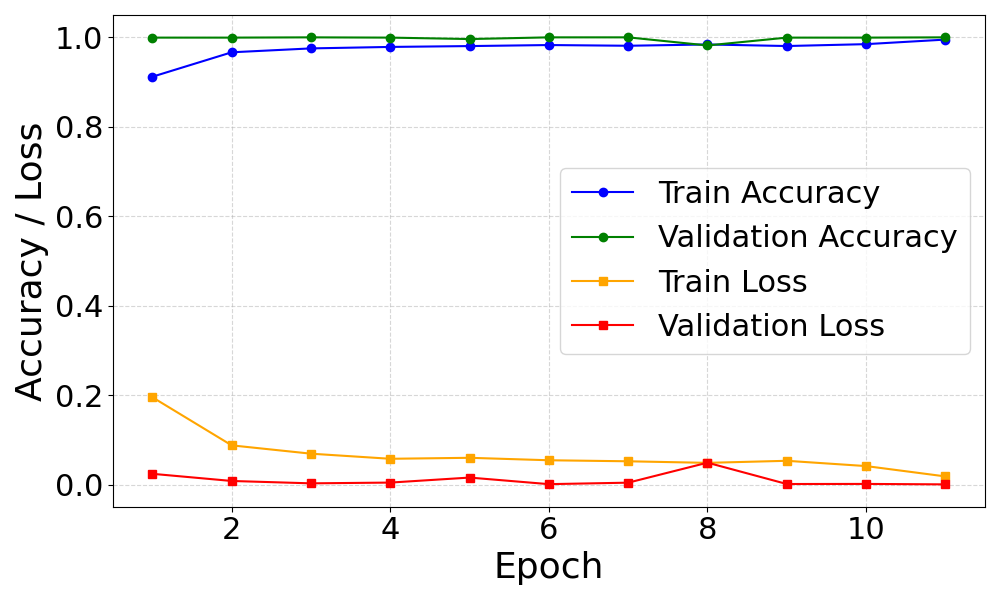}
        \caption{Colon cancer dataset}
        \label{fig:graph_AccVsLoss_colon}
    \end{subfigure}
    \begin{subfigure}[b]{0.36\textwidth}
        \centering
        \includegraphics[width=\textwidth]{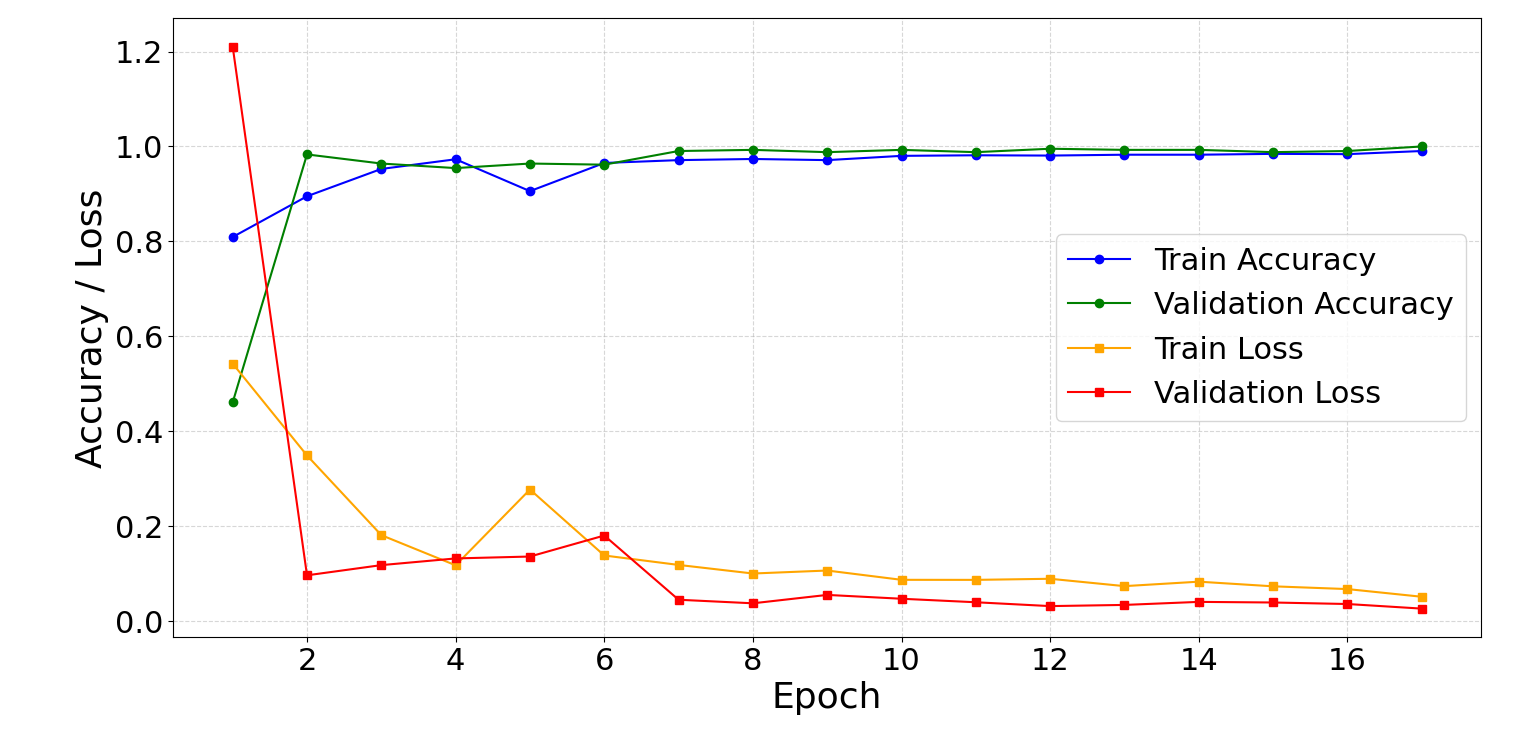}
        \caption{ALL dataset}
        \label{fig:graph_AccVsLoss_ALL}
    \end{subfigure}
    \caption{Training and validation accuracy and loss curves across datasets, demonstrating stable convergence and consistent generalisation performance.}
    \label{fig:graph_AccVsLoss_lung_colon}
\end{figure}

Training and validation accuracy and loss curves are shown in Fig.~\ref{fig:graph_AccVsLoss_lung_colon}. These curves demonstrate stable optimisation and consistent convergence behaviour across datasets. The close agreement between training and validation accuracy indicates effective generalisation, while the decreasing loss curves confirm stable learning without evidence of significant overfitting.

\subsection{Attention Visualisation}
\label{subsec:attention_visualisation}
An important characteristic of Vision Transformer architectures is the self-attention mechanism, which enables identification of spatial regions that contribute most strongly to the classification decision. Unlike convolutional neural networks, where interpretability often requires post hoc methods such as Grad-CAM, the transformer architecture inherently provides attention weights that can be analysed to understand the model’s focus during inference.

\begin{figure}[t]
    \centering
    \begin{subfigure}[b]{0.48\textwidth}
        \centering
        \includegraphics[width=\textwidth]{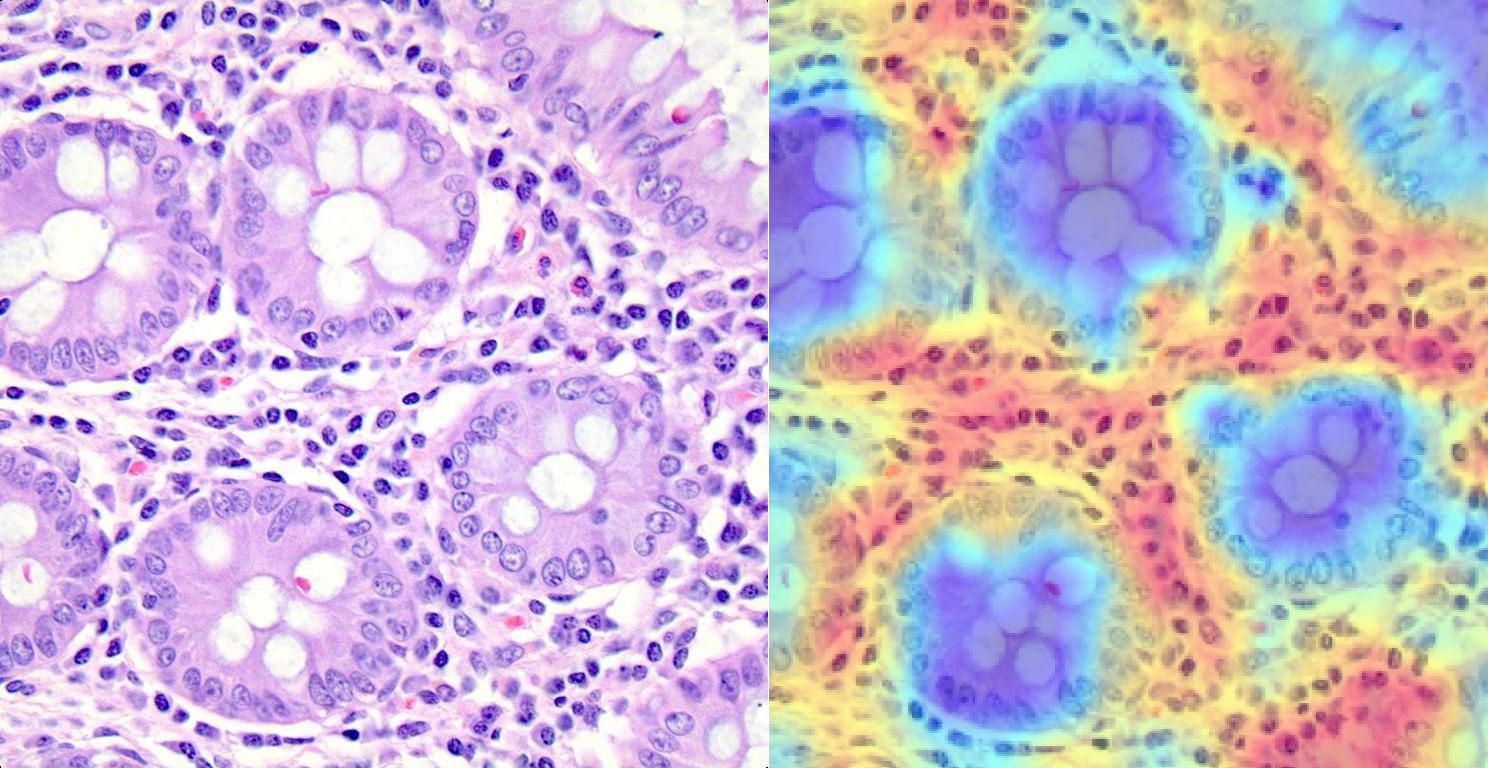}
        \caption{Colon benign tissue (LC25000)}
        \label{fig:Attention_colon}
    \end{subfigure}
    \hfill
    \begin{subfigure}[b]{0.48\textwidth}
        \centering
        \includegraphics[width=\textwidth]{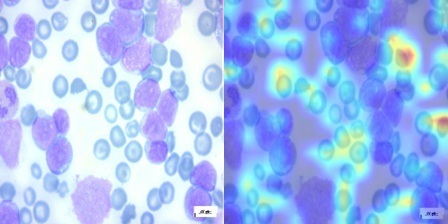}
        \caption{Acute lymphoblastic leukaemia (ALL)}
        \label{fig:Attention_all}
    \end{subfigure}
    \caption{Attention visualisation generated by the proposed DeepHistoViT model. For each example, the original histopathology image is shown alongside the corresponding attention map, highlighting regions that contribute most strongly to the classification decision. The attention maps demonstrate spatial localisation of diagnostically relevant morphological features.}
    \label{fig:Attention_visualisation}
\end{figure}

To examine the spatial regions influencing the model’s predictions, attention maps are extracted from the final transformer encoder layer and projected onto the original input image. These attention maps provide a visual representation of the relative importance assigned to different image regions during classification. Representative examples are shown in Fig.~\ref{fig:Attention_visualisation} for colon tissue and acute lymphoblastic leukaemia samples. In each case, the original histopathology image is shown alongside the corresponding attention map generated by the model. The attention maps highlight spatially localised regions corresponding to diagnostically relevant morphological structures, including cellular arrangements and nuclear features. This behaviour is consistent with the expectation that classification decisions are driven by discriminative tissue patterns. These visualisations provide qualitative evidence that the proposed DeepHistoViT framework learns spatially meaningful feature representations and focuses on relevant tissue regions when performing classification. Such interpretability is particularly important in medical imaging applications, where transparency of model decision-making is essential for clinical reliability and validation.

\subsection{Comparative Analysis with State-of-the-Art Models}
\label{subsec:comparison}

To assess the performance of the proposed DeepHistoViT framework in the context of existing approaches, a comparative analysis is conducted against recently published deep learning models evaluated on the same datasets. This comparison provides insight into the relative performance of the proposed method and its effectiveness for histopathology image classification.


Table~\ref{tab:model_comparison_lung} presents a comparison of classification performance on the lung cancer subset of the LC25000 dataset~\cite{LC25000}. The proposed DeepHistoViT model achieves an overall accuracy of 100\%, with corresponding precision, recall, F1-score, and ROC-AUC values of 100\%. These results are consistent with the performance reported by Shahadat et al.~\cite{Shahadat2025}, who also achieved perfect classification using a deep learning model on the same dataset. Other recent approaches, including ML3CNet proposed by Kumar et al.~\cite{KUMAR2024108207} and the multi-CNN framework proposed by Attallah et al.~\cite{Attallah2025}, report accuracies of 99.72\% and 99.78\%, respectively. These findings indicate that the proposed DeepHistoViT framework achieves performance comparable to the highest reported results on this dataset.

\renewcommand{\arraystretch}{1.3}
\begin{table}
\tiny
\centering
\caption{Comparison of classification performance for lung cancer (LC25000) dataset across recent studies and the proposed DeepHistoViT model.}
\resizebox{\textwidth}{!}{%
\begin{tabular}{|p{3cm}|p{2.5cm}|p{1.5cm}|p{1.5cm}|p{1.5cm}|p{1.5cm}|p{1.5cm}|}
\hline
\textbf{Study} & \textbf{Model} & \textbf{Accuracy (\%)} & \textbf{Precision (\%)} & \textbf{Recall (\%)} & \textbf{F1-Score (\%)} & \textbf{ROC-AUC} \\
\hline
Kumar et al. \cite{KUMAR2024108207} (2024) & ML3CNet & 99.72 & 99.64 & 99.66 & 99.65 & 99.78 \\
\hline
Attallah et al. \cite{Attallah2025} (2025) & Multi-CNN & 99.78 & 99.78 & 99.78 & 99.78 & -- \\
\hline
Shahadat et al. \cite{Shahadat2025} (2025) & DL & 100.00 & 100.00 & 100.00 & 100.00 & 100.00 \\
\hline
\textbf{Proposed} & \textbf{DeepHistoViT} & \textbf{100.00} & \textbf{100.00} & \textbf{100.00} & \textbf{100.00} & \textbf{100.00} \\
\hline
\end{tabular}%
}
\label{tab:model_comparison_lung}
\end{table}


A similar comparison is performed for the colon cancer subset of the LC25000 dataset. As shown in Table~\ref{tab:model_comparison_colon}, the proposed DeepHistoViT model achieves 100\% accuracy across all evaluation metrics. This performance is consistent with the results reported by Shahadat et al.~\cite{Shahadat2025}. Other recent methods, including the multi-CNN approach proposed by Attallah et al.~\cite{Attallah2025} and the EfficientNet-B3 model proposed by Ochoa-Ornelas et al.~\cite{Ochoa-Ornelas2025}, report accuracies of 99.78\% and 99.39\%, respectively. These results demonstrate that the proposed transformer-based architecture achieves performance comparable to or exceeding existing approaches evaluated on this dataset.

\renewcommand{\arraystretch}{1.3}
\begin{table}
\tiny
\centering
\caption{Comparison of classification performance for colon cancer (LC25000) dataset across recent studies and the proposed DeepHistoViT model.}
\resizebox{\textwidth}{!}{%
\begin{tabular}{|p{3cm}|p{2.5cm}|p{1.5cm}|p{1.5cm}|p{1.5cm}|p{1.5cm}|p{1.5cm}|}
\hline
\textbf{Study} & \textbf{Model} & \textbf{Accuracy (\%)} & \textbf{Precision (\%)} & \textbf{Recall (\%)} & \textbf{F1-Score (\%)} & \textbf{ROC-AUC} \\
\hline
Ochoa-Ornelas et al. \cite{Ochoa-Ornelas2025} (2025) & EfficientNet-B3 & 99.39 & 99.39 & 99.39 & 99.39 & -- \\
\hline
Attallah et al. \cite{Attallah2025} (2025) & Multi-CNN & 99.78 & 99.78 & 99.78 & 99.78 & -- \\
\hline
Shahadat et al. \cite{Shahadat2025} (2025) & DL & 100.00 & 100.00 & 100.00 & 100.00 & 100.00 \\
\hline
\textbf{Proposed} & \textbf{DeepHistoViT} & \textbf{100.00} & \textbf{100.00} & \textbf{100.00} & \textbf{100.00} & \textbf{100.00} \\
\hline
\end{tabular}%
}
\label{tab:model_comparison_colon}
\end{table}


The proposed DeepHistoViT framework also demonstrates strong performance on the acute lymphoblastic leukaemia dataset. As shown in Table~\ref{tab:model_comparison_all}, the model achieves an accuracy of 99.85\%, with corresponding precision, recall, and F1-score values of 99.85\%, and an ROC-AUC of 99.99\%. This performance exceeds the results reported by recent studies, including ALL-Net proposed by Thiriveedhi et al.~\cite{Thiriveedhi2025}, which achieved 99.32\% accuracy, and DarkNet19 ESA proposed by Basaran et al.~\cite{BASARAN2025}, which achieved 98.52\% accuracy. These findings indicate that the proposed transformer-based framework achieves competitive performance across multiple histopathology datasets.

\renewcommand{\arraystretch}{1.3}
\begin{table}
\tiny
\centering
\caption{Performance comparison of recent models on the acute lymphoblastic leukaemia dataset.}
\resizebox{\textwidth}{!}{%
\begin{tabular}{|p{3cm}|p{2.5cm}|p{1.5cm}|p{1.5cm}|p{1.5cm}|p{1.5cm}|p{1.5cm}|}
\hline
\textbf{Study} & \textbf{Model} & \textbf{Accuracy (\%)} & \textbf{Precision (\%)} & \textbf{Recall (\%)} & \textbf{F1-Score (\%)} & \textbf{ROC-AUC} \\
\hline
Muhammad et al. \cite{Muhammad2025} (2025) & EfficientNet-B7 & 96.78 & 97.00 & 96.25 & 96.57 & 99.85 \\
\hline
Basaran et al. \cite{BASARAN2025} (2025) & DarkNet19 ESA & 98.52 & 98.40 & 98.22 & 98.30 & 99.95 \\
\hline
Thiriveedhi et al. \cite{Thiriveedhi2025} (2025) & ALL-Net & 99.32 & 99.25 & 99.50 & 99.50 & -- \\
\hline
\textbf{Proposed} & \textbf{DeepHistoViT} & \textbf{99.85} & \textbf{99.85} & \textbf{99.85} & \textbf{99.85} & \textbf{99.99} \\
\hline
\end{tabular}%
}
\label{tab:model_comparison_all}
\end{table}

\section{Discussion}
\label{sec:Discussion}
This work evaluates the effectiveness of a transformer-based architecture, DeepHistoViT, for automated histopathology image classification across multiple cancer types. By combining a pretrained Vision Transformer backbone with a customised multi-layer classification head and selective fine-tuning strategy, the proposed framework achieves high classification performance across all evaluated datasets. In particular, the model attains perfect classification accuracy on the lung and colon cancer subsets of the LC25000 dataset and near-perfect performance on the acute lymphoblastic leukaemia dataset. These findings indicate that transformer-based representations, when appropriately adapted, can effectively capture morphological features relevant for histopathological classification.

The observed perfect classification performance on the LC25000 dataset requires careful interpretation. Previous studies have noted that LC25000 is derived from a limited number of original whole-slide images, which are expanded through extensive augmentation~\cite{LC25000}. As a result, augmented samples may share substantial visual similarity, potentially simplifying the classification task. Consequently, while the reported results demonstrate the model’s ability to learn discriminative features within this dataset, they may not directly translate to real-world clinical scenarios characterised by greater inter-patient and inter-institutional variability. Nevertheless, the strong performance observed on independent datasets such as the ALL dataset suggests that the proposed architecture is capable of generalising beyond a single data source.

The architectural modifications introduced in DeepHistoViT contribute to its observed performance. The proposed multi-layer classification head enables progressive refinement of the high-dimensional feature representations produced by the transformer encoder. By incorporating batch normalisation, non-linear activation, and dropout-based regularisation, this component improves feature discrimination while reducing overfitting, particularly in datasets with limited training samples. In addition, the selective fine-tuning strategy, in which only the final transformer encoder layers are updated during training, allows domain-specific adaptation while preserving general visual representations learned during large-scale pretraining. This approach stabilises optimisation and improves generalisation across datasets with differing staining patterns and imaging characteristics.

Differences in classification performance across datasets highlight the influence of dataset size, diversity, and complexity on model performance. The slightly lower accuracy observed on the ALL dataset compared to LC25000 reflects greater morphological variability and reduced sample size. These findings emphasise the importance of evaluating deep learning models on multiple datasets to obtain a more comprehensive assessment of generalisation performance. Future work may benefit from incorporating larger and more diverse datasets, as well as domain adaptation techniques designed to improve robustness to variations in staining protocols and imaging conditions.

Attention visualisation provides additional insight into the model’s decision-making process. The attention maps generated by DeepHistoViT highlight spatial regions that contribute most strongly to the classification outcome, and these regions correspond to relevant tissue structures and cellular morphology. While these visualisations provide qualitative evidence of meaningful feature utilisation, further quantitative evaluation of interpretability and reliability would be required to support clinical deployment.

Despite the promising results, limitations should be considered. First, the evaluation is conducted on publicly available datasets that may not fully reflect the variability encountered in clinical practice. Second, although the model demonstrates strong performance across multiple datasets, validation on large-scale, multi-centre clinical cohorts would be necessary to establish clinical robustness. Third, while attention mechanisms provide interpretability, additional approaches such as uncertainty estimation and calibration analysis may be required to ensure reliable deployment in safety-critical clinical settings.

Overall, the results demonstrate that transformer-based architectures, combined with appropriate domain adaptation and classification head design, provide a promising approach for histopathological image classification. These findings support further investigation of transformer-based methods for medical image analysis, particularly in scenarios requiring modelling of complex spatial dependencies.

\section{Conclusions}
\label{sec:Conclusions}
This paper presents DeepHistoViT, a Vision Transformer-based framework for histopathology image classification across multiple cancer types. By combining a pretrained ViT backbone with a customised classification head and selective fine-tuning strategy, the proposed model achieves high classification performance across all evaluated datasets, including perfect accuracy on the LC25000 lung and colon datasets and near-perfect performance on the acute lymphoblastic leukaemia dataset. These results demonstrate the effectiveness of transformer-based representations for capturing discriminative histopathological features. In addition to strong quantitative performance, attention-based visualisation provides qualitative insight into the spatial regions contributing to model predictions, supporting interpretability of the learned representations. The ability of the proposed framework to generalise across datasets without explicit stain normalisation further highlights its robustness. Future work will focus on validation using larger, multi-centre clinical datasets and the integration of uncertainty estimation and calibration methods to improve reliability in clinical settings.

\section*{Ethics statement}

This study did not involve the collection of new data from human participants or animals. All datasets used in this work are publicly available and were originally collected and released by their respective providers in accordance with relevant ethical guidelines and institutional regulations. The authors did not have access to any personally identifiable information, and therefore no additional ethical approval or informed consent was required for this secondary analysis.

\bibliographystyle{plainnat}

\bibliography{sample}

\end{document}